\documentclass[10pt,twocolumn,letterpaper]{article}

\usepackage{cvpr2025}      
\usepackage{times}
\usepackage{xcolor}
\usepackage{lipsum}
\usepackage{color, colortbl}
\usepackage{xspace}
\usepackage{capt-of}
\usepackage{wrapfig}
\usepackage{pgfplots}
\pgfplotsset{compat=1.17}
\usepackage{tikz-cd}
\usepackage{amsmath,amssymb,amsfonts,dsfont,pifont,bm,bbm,mathrsfs,mathtools,nicefrac}
\usepackage{algorithm,algpseudocode,listings}
\usepackage{booktabs,multirow,adjustbox,diagbox,threeparttable}
\definecolor{cvprblue}{rgb}{0.21,0.49,0.74}
\usepackage[pagebackref,breaklinks,colorlinks,citecolor=cvprblue]{hyperref}
\usepackage[capitalize]{cleveref}  
\usepackage{capt-of}
\usepackage{wrapfig}
\usepackage{tikz-cd}

\crefname{section}{Sec.}{Secs.}
\Crefname{section}{Section}{Sections}
\crefname{table}{Tab.}{Tabs.}
\Crefname{table}{Table}{Tables}
\crefname{figure}{Fig.}{Figs.}
\Crefname{figure}{Figure}{Figures}
\crefname{equation}{Eq.}{Eqs.}
\Crefname{equation}{Equation}{Equations}
\newcommand{\method}{{\sc flare}\xspace}
\newcommand{\supp}{\textit{supplementary materials}\xspace}

\def\confName{CVPR}
\def\confYear{2025}

\title{FLARE: Feed-forward Geometry, Appearance and Camera Estimation from Uncalibrated Sparse Views }

\author{
Shangzhan Zhang\textsuperscript{1,2*} \and
Jianyuan Wang\textsuperscript{3*} \and
Yinghao Xu\textsuperscript{4*$^\dagger$} \and
Nan Xue\textsuperscript{2} \and
Christian Rupprecht\textsuperscript{3} \and
Xiaowei Zhou\textsuperscript{1\textdagger} \and 
Yujun Shen\textsuperscript{2} \and 
Gordon Wetzstein\textsuperscript{4}
\and 
\textsuperscript{1}Zhejiang University \and
\textsuperscript{2}Ant Group \and
\textsuperscript{3}University of Oxford \and
\textsuperscript{4}Stanford University
}
\newcommand\blfootnote[1]{%
  \begingroup
  \renewcommand\thefootnote{}\footnote{#1}%
  \addtocounter{footnote}{-1}%
  \endgroup
}
\begin{document}
\twocolumn[{%
\renewcommand\twocolumn[1][]{#1}%
\maketitle
\centering
\includegraphics[width=\textwidth]{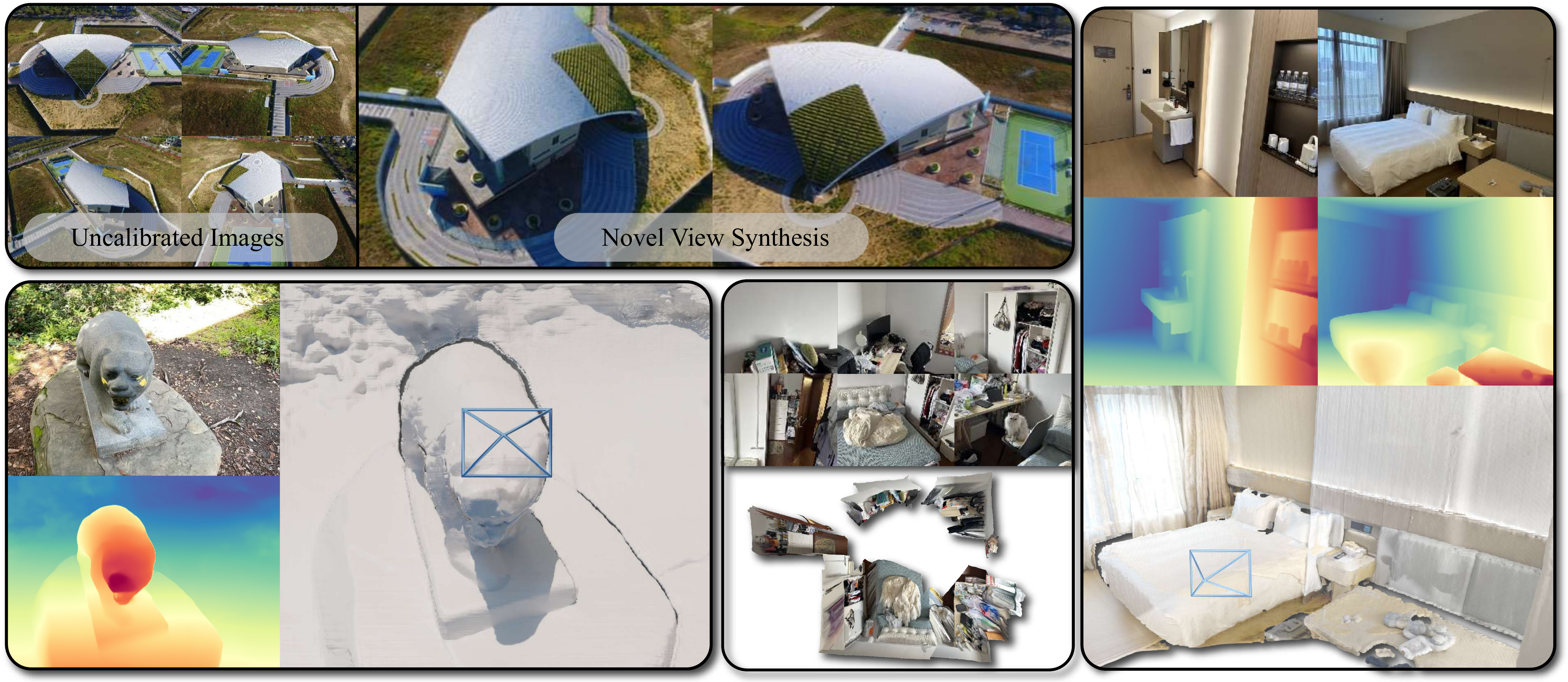}

\captionof{figure}{We present \method, a feed-forward approach that simultaneously recovers high-quality poses, geometry and appearance from uncalibrated sparse views within \textbf{0.5s}.
Our model excels in the scenarios with a camera circling around the subject, and also shows robust generalization to real-world casual captures, such as an indoor bedroom.
In the central area below, we casually captured six random bedroom images with minimal overlap. Our method demonstrates high-quality geometry reconstruction even in this challenging case.
}
\vspace{1em}
\label{fig:teaser}}]

\blfootnote{$^*$The first three authors contributed equally. $^\dagger$Corresponding author.}

\begin{abstract}
We present \method, a feed-forward model designed to infer high-quality camera poses and 3D geometry from uncalibrated sparse-view images (\textit{i.e.}, as few as 2-8 inputs), which is a challenging yet practical setting in real-world applications.
Our solution features a cascaded learning paradigm with camera pose serving as the critical bridge, recognizing its essential role in mapping 3D structures onto 2D image planes.
Concretely, \method starts with camera pose estimation, whose results condition the subsequent learning of geometric structure and appearance, optimized through the objectives of geometry reconstruction and novel-view synthesis.
Utilizing large-scale public datasets for training, our method delivers state-of-the-art performance in the tasks of pose estimation, geometry reconstruction, and novel view synthesis, while maintaining the inference efficiency (\textit{i.e.}, less than 0.5 seconds). The project page and code can be found at: \url{https://zhanghe3z.github.io/FLARE/}

\end{abstract}
\section{Introduction}\label{sec::intro}
Reconstructing 3D scenes from multi-view images is a fundamental problem with wide-ranging applications across computer vision, perception, and computer graphics. 
Traditional approaches typically solve this problem in two stages: first, estimating camera parameters using Structure-from-Motion (SfM) solvers~\citep{hartley2003multiple, schonberger2016structure, snavely2006photo}, and then predicting dense depth maps by Multi-View Stereo (MVS) to achieve dense 3D reconstruction~\citep{seitz2006comparison, schonberger2016pixelwise}. 
Despite the significant success of the SfM-MVS paradigm over the past decades, it faces two key limitations. First, these methods rely heavily on handcrafted feature matching and are non-differentiable, preventing them from fully leveraging recent advancements in deep learning. Second, traditional SfM approaches struggle with sparse views and limited viewpoints, significantly restricting their applicability in real-world scenarios.

Recent efforts to tackle these issues have shown potential, but significant challenges remain.  Optimization-based approaches like BARF~\citep{lin2021barf} and NeRF\texttt{-}\texttt{-}\citep{wang2021nerf} jointly optimize camera poses and geometry, but they require a good initialization and suffer from poor generalization to novel scenes. 
Deep camera estimation methods~\citep{sinha2023sparsepose, wang2023posediffusion, lin2023relposepp, zhang2024cameras, Rockwell2024, Rockwell2022} treat sparse-view SfM as a camera parameter regression problem, yet struggle with accuracy and generalization. VGGSfM~\citep{wang2024vggsfm} improves this by incorporating multi-view tracking and differentiable bundle adjustment but falls short in providing dense geometry.
DUSt3R\citep{wang2024dust3r} and MASt3R~\citep{leroy2024grounding} propose generating a two-view point map with pixel-wise geometry, but their reliance on post-optimization global registration is time consuming and often yields suboptimal results. PF-LRM~\citep{wang2023pf} offers feed-forward reconstruction from four images, but its tri-plane representation~\cite{chan2022efficient} limits performance in large-scale scenes.
While these methods have demonstrated promising advances in sparse-view settings, they still lack a comprehensive solution that combines scalability, accuracy, and efficiency in camera, geometry, and appearance estimation.
     

We present \method, a novel feed-forward and differentiable system that infers high-quality geometry, appearance, and camera parameters from uncalibrated sparse-view images. 
Direct optimization of these parameters from images often presents significant learning difficulties, frequently converging to sub-optimal solutions with distorted geometry and blurry textures. To address these challenges, we introduce a novel cascade learning paradigm that progressively estimates camera poses, geometry, and appearance, relaxing traditional requirements for 3D reconstruction such as dense image views, accurate camera poses, and wide baselines.
Our cascade learning paradigm lies in decomposing the challenging  optimization problem into sequential stages, using camera poses as proxies for each stage. 
The central concept is that a camera pose frames a 2D image within a 3D observation frustum, reducing the learning complexity for subsequent tasks. 
Our method starts with a neural pose predictor that estimates coarse camera poses from sparse-view images. These initial poses provide geometric cues that facilitate a transformer-based architecture to refine the poses, compute point maps, and predict 3D Gaussians for novel-view synthesis. 
For geometry prediction, we introduce a two-stage approach instead of direct global prediction. First, we estimate camera-centric point maps in individual camera coordinates. Then, a neural scene projector unifies these local point maps into a coherent global structure. With this approach, we enable faster convergence in geometry learning and reduce geometric distortion for challenging scenes.

We trained our model on a set of large public datasets~\cite{li2018megadepth, baruch2021arkitscenes, yao2020blendedmvs, yeshwanth2023scannet++, Reizenstein2021CommonOI, Sun_2020_CVPR, xia2024rgbd, Ling_2024_CVPR} . \method achieves state-of-the-art results in camera pose estimation, point cloud estimation, and novel-view synthesis. With unposed images as input, \method can produce photorealistic novel-view synthesis using Gaussian Splatting in just 0.5 seconds, which is a substantial improvement over previous optimization-based methods. 
As demonstrated in  \cref{fig:teaser}, our system reconstructs 3D scenes and estimates poses from as few as 2-8 input images.

The primary contributions of this work are as follows:
\begin{itemize}[leftmargin=1.5em]
\item We propose an efficient, feed-forward, and differentiable system for high-quality 3D Gaussian scene reconstruction from uncalibrated sparse-view images, achieving inference in less than 0.5 seconds.

\item We demonstrate that leveraging camera poses as proxies effectively simplifies complex 3D learning tasks. We thus introduce a novel cascaded learning paradigm that starts with camera pose estimation, whose results condition the subsequent learning of geometric structure and appearance. 

\item We propose a two-stage geometry learning approach that first learns camera-centric point maps and builds a global geometry
projector to unify the point maps into a global coordinate.

\end{itemize}

\section{Related Work}\label{sec::related}

\noindent\textbf{Structure-from-Motion (SfM).} SfM techniques aim to estimate camera poses and reconstruct sparse 3D structures.
Conventional approaches~\citep{schonberger2016structure, snavely2006photo} employ multi-stage optimization, starting with pairwise feature matching~\citep{matas2004robust, bay2006surf, lowe2004distinctive} across views to establish correspondences, followed by camera pose optimization using incremental bundle adjustment~\citep{triggs2000bundle}.
Recently, numerous learning-based SfM methods have been proposed to enhance the traditional multi-stage pipeline. These improvements focus on three main areas: developing learning-based feature descriptors~\citep{detone2018superpoint, dusmanu2019d2, yi2016lift}, learning more accurate matching algorithms~\citep{sun2021loftr, sarlin2020superglue}, and implementing differentiable bundle adjustment~\citep{wang2023visual, lin2021barf}.
However, when input views are extremely sparse, accurately matching features becomes highly challenging, leading to degraded camera pose estimation performance.

\noindent\textbf{Multi-view Stereo (MVS).}
MVS techniques aim to reconstruct dense 3D geometry from multiple calibrated images. Traditional MVS methods~\citep{galliani2015massively, schonberger2016pixelwise, furukawa2015multi} typically follow a pipeline of depth map estimation, depth map fusion, and surface reconstruction. These approaches often rely on photometric consistency across views and various regularization techniques~\citep{yao2019recurrent} to handle challenging scenarios.
Recent years have witnessed a surge in learning-based MVS methods, leveraging deep neural networks to improve reconstruction quality and efficiency either with cascade cost volume matching~\citep{yao2018mvsnet, yao2019recurrent} or reconstruction supervision with differentiable rendering~\citep{chen2021mvsnerf, chen2024mvsplat, charatan2024pixelsplat, xu2024grm, hong2023lrm, li2023instant3d}.
Despite significant progress, MVS methods consistently depend on calibrated camera poses, which are typically estimated by SfM methods. This cascaded pipeline often causes MVS to perform suboptimally when the estimated poses are inaccurate.
DUSt3R~\citep{wang2024dust3r, leroy2024grounding} directly predicts the geometry of visible surfaces without any explicit knowledge of the camera parameters. However, under multi-view settings, their approach is limited to pairwise image processing followed by global alignment, failing to fully exploit multi-view information and supporting photorealistic rendering.

\noindent\textbf{3D Reconstruction from Sparse-view Images.}  Neural representations~\citep{occupancy,deepsdf,nerf,sitzmann2019scene,tewari2022advances} present a promising foundation for scene representation and neural rendering.
When applied to novel-view synthesis, these methods have demonstrated success in scenarios with dense-view training images, showcasing proficiency in single-scene overfitting.
Notably, recent advancements~\citep{pixelnerf, mvsnerf, sparseneus, ibrnet, visionnerf, dietnerf, chen2024mvsplat, zhang2024worldconsistentvideodiffusionexplicit} have extended these techniques to operate with a sparse set of views, displaying improved generalization to unseen scenes.
These methods face challenges in capturing multiple modes within large-scale datasets, resulting in a limitation to generate realistic results.
Additional works~\citep{xu2024grm, li2023instant3d, hong2023lrm, charatan2024pixelsplat} further scale up the model and datasets for better generalization with NeRF or Gaussian Splatting.
Unlike existing methods that rely on calibrated camera poses to supervise the neural network training, our approach can perform direct 3D reconstruction from uncalibrated images.

\noindent\textbf{Pose-free Novel-view Synthesis.}
Recent research has made significant progress in novel-view synthesis from uncalibrated images.
One line of research focuses on jointly optimizing camera poses and radiance fields from dense-view images. BARF~\cite{lin2021barf}, NeRF\texttt{-}\texttt{-}~\cite{wang2021nerf}, and subsequent works~\cite{Jeong_2021_ICCV, bian2023nope, truong2023sparf} have advanced this approach.
Several recent methods~\cite{Fu_2024_CVPR, keetha2024splatam, fan2024instantsplat} have extended the 3D representation from NeRF to 3D Gaussians.
Another research direction~\cite{Sajjadi2022RUSTLN, kani24upfusion, xu2025sparp, jiang2024forge} focuses on developing feed-forward novel-view synthesis for unposed images. 
SRT~\cite{srt22} proposes the first pose- and geometry-free framework for novel view synthesis, while LEAP~\cite{jiang2022LEAP} pioneers  pose-free radiance field reconstruction by directly estimating scene geometry and radiance fields. 
FlowCam~\cite{smith2023flowcam} and FlowMap~\cite{smith2024flowmap} introduce 2D flow to enable unsupervised learning of generalizable 3D reconstruction, though their performance degrades in sparse-view settings.
PF-LRM~\cite{wang2023pf} estimates camera poses by predicting point maps and solving a differentiable perspective-n-point (PnP) problem, but shows limited generalization to complex 3D scenes.
PF3plat~\cite{hong2024pf3plat} achieves coarse alignment of 3D Gaussians by leveraging pre-trained models for monocular depth estimation and visual correspondence.
Splatt3R~\cite{smart2024splatt3r} and NopoSplat~\cite{ye2024no} utilize DUSt3R~\cite{wang2024dust3r} or MASt3R~\cite{leroy2024grounding} to predict point maps as proxy geometry and subsequently learn 3D Gaussians for sparse-view reconstruction.
However, existing approaches are either restricted to two-view scenarios or produce suboptimal rendering results due to imperfect geometry estimates from DUSt3R or MASt3R.
Our work presents a differentiable system that simultaneously predicts camera parameters, geometry, and appearance, achieving superior generalization across diverse real-world scenes.

\section{Method}\label{sec::method}

\begin{figure*}[htbp]

\begin{center}
\includegraphics[width=\textwidth]{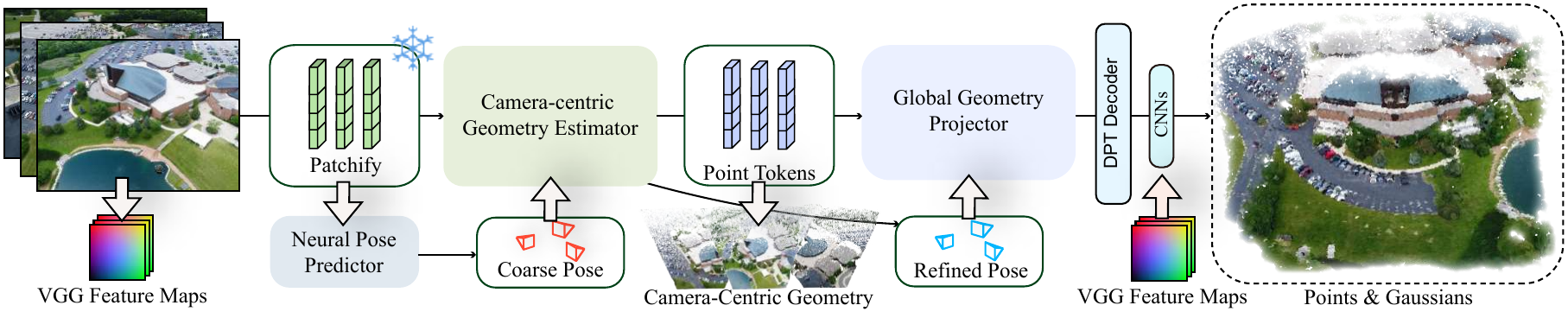}

\end{center}
\vspace{-1em}
\caption{
    \textbf{Illustration of our pipeline.}
    Given  uncalibrated sparse views, our model can infer high-quality camera poses, geometry and appearance in a single feed-forward pass. We use camera poses as proxies to guide subsequent geometry and appearance learning. Given initial pose estimates, we first compute camera-centric geometry, then project it into a global scene representation. Finally, we form 3D Gaussians on top of the scene geometry to enable photo-realistic novel-view synthesis. 
}
\label{fig:pipeline}
\vspace{-1em}

\end{figure*}


\method uses point maps~\cite{wang2024dust3r} as our geometry representation for two key advantages: their compatibility with neural networks and natural integration with 3D Gaussians for appearance modeling.
As shown in \cref{fig:pipeline}, \method is  feed-forward model to infer high-quality camera poses and 3D geometry from uncalibrated sparse-view images.
Our solution is a cascaded learning paradigm that first estimates camera poses from sparse views and then leverages the estimates to guide the subsequent geometry and appearance learning.
%
We present our neural pose predictor for sparse-view pose estimation in \cref{method:pose}.
With the pose estimates, we propose a two-stage learning paradigm for geometry estimation (\cref{method:geometry}). Given the estimated geometry, we develop a 3D Gaussian reconstruction head that enables high-quality appearance modeling for photorealistic novel view synthesis (\cref{method:apperance}). Finally, we detail the training objectives for the whole framework (\cref{method:training}). 

\subsection{Neural Pose Predictor}
\label{method:pose}
Traditional pose estimation methods rely on feature matching to find correspondences, but this often fails in sparse-view scenarios where images have limited overlapping regions. 
Inspired by previous deep camera estimation methods~\cite{wang2023posediffusion, wang2024vggsfm}, we drop the feature matching and formulate pose estimation as a direct transformation problem from image space to camera space using an end-to-end transformer model. Given the input images $\mathcal{I} = \{\mathbf{I}_i\}_{i=1}^{N}$,  we first tokenize them into non-overlapping patches to obtain image tokens. We then initialize learnable camera latents $\mathcal{Q}_{c} = \{\mathbf{q}^{\text{coarse}}_i\}_{i=1}^{N}$. By concatenating the image patches with camera latents into a 1D sequence, we leverage a small decoder-only transformer, named ``neural pose predcitor'' $\mathrm{F}_p(\cdot)$ to estimate the coarse camera poses $\mathcal{P}_{c} = \{\mathbf{p}^{\text{coarse}}_i\}_{i=1}^{N}$:
\begin{equation}
\mathcal{P}_{c} = \mathrm{F}_p(\mathcal{Q}_{c}, \mathcal{I}).
\end{equation}
We parametrize each camera pose as a 7-dimensional vector comprising the absolute translation and a normalized quaternion. 
With this neural pose predictor, we can estimate coarse camera poses as initialization for subsequent geometry and appearance learning. 
We observe that the estimated poses do not need to be very accurate---only approximating the ground truth distribution is enough. This aligns with our key insight: camera poses, even imperfect, provide essential geometric priors and spatial initialization, which significantly reduces the complexity for geometry and appearance reconstruction. 

\subsection{Multi-view Geometry Estimation}\label{method:geometry}
With our estimated camera poses serving as an effective intermediate representation, we propose a two-stage geometry learning approach to improve reconstruction quality. 
Our key idea is to first learn camera-centric geometry in local frames (camera coordinate system) and then build a neural scene projector to transform it into a global world coordinate system with the guidance of estimated poses. 

\noindent\textbf{Camera-centric Geometry Estimation.} 
Learning geometry in camera space aligns with the image formation process, as each view directly observes local geometry from its perspective. This also simplifies the geometry learning by focusing on local structures visible in each view, rather than reasoning about complex global spatial relationships.
We tokenize images into image tokens and concatenate them with camera tokens derived from coarse pose estimates $\mathcal{P}_c$. These tokens are then fed into a transformer architecture $\mathrm{F}_l(\cdot)$ to estimate local point tokens $\mathcal{T}_l = \{\mathbf{T}^{local}_{i}\}_{i=1}^{N}$.
The self-attention mechanism in the transformer can perform association across different views and exploit the geometric cues from cameras.
The local point tokens are subsequently fed into a DPT-based~\cite{ranftl2021vision} decoder $\mathrm{D}_l(\cdot)$ for spatial upsampling to obtain dense point maps $\mathcal{G}_l = \{\mathbf{G}^{\text{local}}_{i}\}_{i=1}^{N}$ and confidence map $\mathcal{C}_l = \{\mathbf{C}^{\text{local}}_{i}\}_{i=1}^{N}$ in local camera space. Meanwhile, we further refine the initial camera poses by introducing additional learnable pose tokens $\mathcal{Q}_{f}$ into the network,  which output refined pose estimates $\mathcal{P}_f$ alongside the geometry prediction:
\begin{align}
\mathcal{T}_{l}, \mathcal{P}_{f} &= \mathrm{F}_l(\mathcal{I}, \mathcal{P}_{c}, \mathcal{Q}_{f}) \\
\mathcal{G}_{l}, \mathcal{C}_{l} &= \mathrm{D}_l(\mathcal{T}_{l}).
\end{align}
We can process an arbitrary number of images, as long as the GPU memory does not overflow. For a detailed explanation, please refer to the \supp.
We find this multi-task learning scheme can boost each task performance by providing complementary supervision signals between pose refinement and geometry estimation, as observed in previous work~\cite{wang2023pf}. To handle potentially inaccurate pose estimates during inference, we introduce a simple yet effective pose augmentation strategy during training. Specifically, we randomly perturb the predicted camera poses by adding Gaussian noise, which allows the network to learn to adapt noisy estimated poses at inference time.


\noindent\textbf{Global Geometry Projection.}
We aim to transform camera-centric geometry predictions into a consistent global geometry using refined camera poses. However, this transformation is challenging since imperfect pose estimates make direct geometric reprojection unreliable.
Rather than using geometric transformation, we propose a learnable geometry projector $\mathrm{F}_g(\cdot)$ that transforms local geometry $\mathcal{G}_l$ into global space, conditioned on the estimated poses $\mathcal{P}_f$. This learned approach is more robust to pose inaccuracies compared to direct geometric projection.
For computational efficiency, we utilize the local point tokens $\mathcal{T}_l$ rather than the dense camera geometry $\mathcal{G}_l$ as the input:
\begin{align}
    \mathcal{T}_g &= \mathrm{F}_g(\mathcal{T}_l, \mathcal{P}_f) \\
    \mathcal{G}_{g}, \mathcal{C}_{g} &= \mathrm{D}_g(\mathcal{T}_{g}).
\end{align}
where $\mathrm{D}_g(\cdot)$, $\mathcal{G}_g$ and $\mathcal{C}_g$  denotes the DPT-based upsampling decoder, global point maps and corresponding confidence map.
This geometry projector $\mathrm{F}_g(\cdot)$ is also implemented with a transformer architecture, which is the same as the $\mathrm{F}_l(\cdot)$ but takes different input.

\subsection{3D Gaussians for Appearance Modeling}\label{method:apperance}

Based on the learned 3D point maps, we initialize 3D Gaussians by using the point maps as the centers of 3D Gaussians. We then build a Gaussian regression head to predict other Gaussian parameters including opacity $\mathcal{O} = \{\mathbf{o}_i\}_{i=1}^N$, rotation $\mathcal{R} = \{\mathbf{r}_i\}_{i=1}^N$, scale $\mathcal{S} = \{\mathbf{s}_i\}_{i=1}^N$ and spherical harmonics coefficient $\mathcal{SH} = \{\mathbf{sh}_i\}_{i=1}^N$ for appearance modeling. Specifically, to efficiently model appearance, we utilize a pretrained VGG network to extract features from input images $\mathcal{V} = \{\mathbf{v}_i\}_{i=1}^N$ and build another DPT head on top of the $\mathrm{F}_g(\cdot)$ to obtain an appearance feature $\mathcal{A}$. This appearance feature is then fused with VGG features and fed into a shallow CNN decoder $\mathrm{F}_a(\cdot)$ for Gaussian parameter regression.

To address scale inconsistency between estimated and ground truth geometry, we normalize both into a unified coordinate space. We compute average scale factors from predicted $s = \mathrm{avg}(\mathcal{G}_g)$ and ground truth $s_{gt} = \mathrm{avg}(\mathcal{G}_{gt})$ point maps, normalizing scenes to unit space during rendering. The Gaussian scale parameter $\mathcal{S}$ and novel-view camera position $\mathbf{p}'$ are also normalized.
We use the differentiable Gaussian rasterizer $\mathrm{R}(\cdot)$ to render images with the normalized 3D Gaussians:
\begin{align}
   \mathbf{I}_{\mathbf{p}'} = \mathrm{R}(\{\mathcal{G}_g/s, \mathcal{O}, \mathcal{R}, \mathcal{S}/s, \mathcal{SH}\}, \mathbf{p}'/s_{gt}),
\end{align}
where $\mathbf{I}_{\mathbf{p}'}$ is the rendered image. The entire rendering process is differentiable, enabling end-to-end optimization of the Gaussian regression head through reconstruction loss.



\subsection{Training Loss}\label{method:training}
Our model is a joint learning framework and trained with a multi-task loss function comprising three components: camera pose loss, geometry loss, and Gaussian splatting loss. 
The camera pose loss is defined as the combined sum of rotation and translation losses following the pose loss used in VGGSFM:
\begin{align}
     \mathcal{L}_{\text{pose}} &= \sum_{i=1}^{N} \ell_{\text{huber}}(\mathbf{p}_i^{\text{coarse}}, \mathbf{p}_i) + \ell_{\text{huber}}(\mathbf{p}_i^{\text{fine}}, \mathbf{p}_i),
\end{align}
where $\mathbf{p}_i$ is the ground-truth camera pose of $i$-th image and $\ell_{\text{huber}}$ is the Huber-loss between the parametrization of poses.

The geometry loss includes a confidence-aware 3D regression term similar to that in DUSt3R:
\begin{align}
        \mathcal{L}_{\text{geo}} &= \sum_{i=1}^{N} \sum_{j\in\mathcal{D}^i} \mathbf{C}_{i, j}^\text{camera}\ell_{\text{regr}}^{\text{camera}}(j,i) - \alpha\log \mathbf{C}_{i, j}^\text{camera} \\ &+ \sum_{i=1}^{N} \sum_{j\in\mathcal{D}^i} \mathbf{C}_{i, j}^\text{global}\ell_{\text{regr}}^{\text{global}}(j,i) - \alpha\log \mathbf{C}_{i, j}^\text{global},
\end{align}
where $\mathcal{D}^i$ denotes the valid pixel grids, $\mathbf{C}_{i, j}^\text{local}$ and $\mathbf{C}_{i, j}^\text{global}$ denote the confidence scores of pixel $j$ of $i$-th image in local and global maps. $\ell_{\text{regr}}^{\text{local}}(j,i)$ and $\ell_{\text{regr}}^{\text{global}}(j,i)$ denote the Euclidean distances of pixel $j$ of $i$-th image between the normalized predicted point maps and ground-truth point maps in camera and global coordinate frames.

The Gaussian splatting loss is computed as the sum of the $L_2$ loss and the VGG perceptual loss $L_\text{perp}$ between the rendered $\mathbf{I}_{\mathbf{p}'}$ and ground truth $\hat{\mathbf{I}}_{\mathbf{p}'}$ images. Additionally, we include a depth loss to supervise  rendered depth maps $\mathbf{D}_{\mathbf{p}'}$ with the prediction $\hat{\mathbf{D}_{\mathbf{p}'}}$ from the monocular depth estimator~\cite{depthanything}:
\begin{align}
    \mathcal{L}_{\text{splat}} &= \sum_{\mathbf{p}' \in \mathcal{P}' }\|\hat{\mathbf{I}}_{\mathbf{p}'} - \mathbf{I}_{\mathbf{p}'}\| + 0.5L_\text{perp}(\hat{\mathbf{I}}_{\mathbf{p}'}, \mathbf{I}_{\mathbf{p}'}) \\
    &+ 0.1 \|(\mathbf{W}\hat{\mathbf{D}_{\mathbf{p}'}} + \mathbf{Q}) - \mathbf{D}_{\mathbf{p}'}\|, 
\end{align}
where $\mathbf{W}$ and $\mathbf{Q}$ are the scale and shift used to align  $\hat{\mathbf{D}_{\mathbf{p}'}}$ and $\mathbf{D}_{\mathbf{p}'}$, $\mathcal{P}'$ is the novel-view camera poses.

The total loss is represented as:
\begin{align}
    \mathcal{L}_{\text{total}} &= \lambda_{\text{pose}} \mathcal{L}_{\text{pose}} + \lambda_{\text{geo}} \mathcal{L}_{\text{geo}} + \lambda_{\text{splat}} \mathcal{L}_{\text{splat}},
\end{align}
where $\lambda_{\text{pose}}$,  $\lambda_{\text{geo}}$ and $\lambda_{\text{splat}}$ denotes the loss weight for corresponding loss.

\section{Experiment}\label{sec::exp}

\paragraph{Datasets.}
Following DUSt3R~\cite{wang2024dust3r} and MASt3R~\cite{leroy2024grounding}, we train our model on a mixture of the following public datasets: MegaDepth~\cite{li2018megadepth}, ARKitScenes~\cite{baruch2021arkitscenes}, Blended MVS~\cite{yao2020blendedmvs}, ScanNet++~\cite{yeshwanth2023scannet++}, CO3D-v2~\cite{Reizenstein2021CommonOI}, Waymo~\cite{Sun_2020_CVPR}, WildRGBD~\cite{xia2024rgbd}, and DL3DV~\cite{Ling_2024_CVPR}. These datasets feature diverse types of scenes.

\begin{table}
\centering
\caption{\textbf{Comparison on multi-view pose estimation.} We compare our method to baselines using the RealEstate10K. We use RRA@5°, RTA@5° and AUC@30° to evaluate the pose accuracy.}
\label{tab:pose} 
\resizebox{0.48\textwidth}{!}{
\begin{tabular}{lccc}
\toprule
Optimized-Based Method & RRA@5° $\uparrow$ & RTA@5° $\uparrow$ & AUC@30° $\uparrow$ \\ 
\midrule
DUSt3R~\cite{wang2024dust3r} & 0.83 & 0.37 &  54.9 \\ 
MASt3R~\cite{leroy2024grounding} & 0.87 & 0.45 & 61.1  \\ 
COLMAP~\cite{schoenberger2016sfm} + SPSG~\cite{sarlin2020superglue} & 0.74 & 0.22 & 33.8  \\ 
COLMAP~\cite{schoenberger2016sfm}  & 0.63 & 0.07 & 16.0 \\ 
PixSfM~\cite{lindenberger2021pixsfm}  & 0.70 & 0.14 & 29.9 \\ 
VGGSfM~\cite{wang2024vggsfm}  & -- & -- & 72.1 \\ 
RelPose~\cite{zhang2022relpose} & 0.47 & 0.32 & 11.1 \\ 
\midrule
Feed-forward Method & RRA@5° $\uparrow$ & RTA@5° $\uparrow$ & AUC@30° $\uparrow$ \\ 
\midrule
{PoseDiffusion}~\cite{wang2023posediffusion} & 0.78 & 0.27 & 51.6 \\ 
Ours & \textbf{0.97}  & \textbf{0.65}  & \textbf{79.9} \\ 
\bottomrule
\end{tabular}
}
\vspace{-1em}
\end{table}

\begin{table*}
\centering
\tabcolsep=0.15cm  
\caption{\textbf{Comparison on Sparse-view Reconstruction.} The evaluation requires models to estimate both camera poses and scene geometry in the challenging sparse-view setting. For scene geometry, we report accuracy, completeness, and overall Chamfer distance. We use the accuracy for AUC under 30$^\circ$ degrees for camera poses.
}
\vspace{-0.5em}
\begin{threeparttable}
\resizebox{0.99\linewidth}{!}{
\begin{tabular}{l c c c c| c c c c | c c c c }
\toprule
\multirow{2}{*}{Methods} & \multicolumn{4}{c|}{DTU} & \multicolumn{4}{c|}{ETH3D} & \multicolumn{4}{c}{TUM RGBD} \\
\cmidrule{2-5}  \cmidrule{6-9}  \cmidrule{10-13}
& ACC. $\downarrow$ & COMP. $\downarrow$ & Overall$\downarrow$ & AUC@30$^\circ$↑ & ACC.$\downarrow$ & COMP.$\downarrow$ & Overall$\downarrow$ & AUC@30$^\circ$↑ & ACC.$\downarrow$ & COMP.$\downarrow$ & Overall$\downarrow$ & AUC@30$^\circ$↑ \\
\midrule
DUSt3R  & 3.8562 & 3.1219 & 3.4891 & 9.5 & 0.4856 &  0.7114 &  0.5984 & 10.0 & 0.3228 & 0.4388 & 0.3808 & 23.0\\
MASt3R & 4.2380 & 3.2695 & 3.7537 & 12.3 & \textbf{0.3417} & \textbf{0.3626} & \textbf{0.3522} & 9.9 & 0.2547 & 0.3074 & 0.2810 & 37.4\\
Spann3R &  4.3097 & 4.5573 & 4.4335 & - & 1.1589 & 0.8005 & 0.9797 & - &  0.5031 & 0.4643 & 0.4837 & -\\
\textbf{Ours}    & \textbf{3.5049} & \textbf{2.7254} & \textbf{3.1152} & \textbf{28.1} & 0.4787 & 0.5169 & 0.4978 & \textbf{15.3} & \textbf{0.2046} & \textbf{0.2050} & \textbf{0.2048} & \textbf{53.6}\\

\bottomrule
\end{tabular}
}
\end{threeparttable}
\label{tab:recon}
\end{table*}

\paragraph{Implementation details.} 
Our model is trained from scratch using 8 views as input, without any pre-trained models, except for the encoder.
The neural pose predictor $\mathrm{F}_{p}(\cdot)$ consists of 12 transformer blocks with channel width 768.
In cascade geometry estimation, our camera-centric geometry estimator $\mathrm{F}_{l}(\cdot)$  and global geometry projector $\mathrm{F}_{g}(\cdot)$ both use 12 transformer blocks with channel width 768.
%
We use the Adam~\cite{kingma2014adam} optimizer with an initial learning rate of $1\times10^{-4}$, gradually decreasing to $1\times10^{-5}$. We train our model on 64 NVIDIA A800 GPUs for 200 epochs with an input resolution of $512\times384$. The training takes approximately 14 days to complete. We use gsplat~\cite{ye2025gsplat} for efficient Gaussian splatting rendering.
%
%
More implementation details are presented in the \supp.

\paragraph{Inference.}
Our model, although trained with 8 views, generalizes well to scenarios ranging from as few as 2 views to as many as 25 views. We give a comprehensive study between performance and view numbers in \supp. 

\subsection{Multi-view Pose Estimation}
\noindent\textbf{Dataset.}
Following PoseDiffusion~\cite{wang2023posediffusion}, we evaluate the camera pose estimates on the RealEstate10K dataset~\cite{zhou2018stereo}. We apply our method directly to the RealEstate10K dataset without fine-tuning, using 5 images as input following previous protocol. 

\noindent\textbf{Metrics.}
We evaluate pose accuracy on the RealEstate10K dataset~\cite{zhou2018stereo} using three metrics~\cite{jin2021image, wang2023posediffusion}: AUC, RRA, and RTA.
The AUC metric computes the area under the accuracy curve across different angular thresholds, where accuracy is determined by comparing the angular differences between predicted and ground-truth camera poses.
RRA (Relative Rotation Accuracy) and RTA (Relative Translation Accuracy) measure the angular differences in rotation and translation respectively. The final accuracy at a threshold $\tau$ is determined by the minimum of RRA@$\tau$ and RTA@$\tau$.

\noindent\textbf{Baseline.}
For camera pose estimation, we compare our method with recent deep optimization-based methods, including DUSt3R~\cite{wang2024dust3r}, MASt3R~\cite{leroy2024grounding}, VGGSfM~\cite{wang2024vggsfm} and RelPose~\cite{zhang2022relpose}. We also include traditional SfM methods such as COLMAP~\cite{schoenberger2016sfm} and PixSfM~\cite{schonberger2016pixelwise}, as well as feed-forward method PoseDiffusion~\cite{wang2023posediffusion}.
%

\noindent\textbf{Comparison.} We show quantitative comparison in \cref{tab:pose}.
Conventional SfM methods like COLMAP and PixSfM rely on time-consuming bundle adjustment, resulting in slow inference speed for camera pose estimation. These optimization-based methods also struggle with sparse views where feature correspondences are difficult to establish.
DUSt3R and MASt3R employ global alignment optimization, which not only leads to slow inference but also limits their performance due to their two-view geometry learning paradigm that cannot effectively leverage multi-view associations.
Compared to other feed-forward approaches, our method achieves superior performance, thanks to the coarse-to-fine pose estimation strategy and multi-task learning with geometry.

\subsection{Sparse-view 3D Reconstruction}
\noindent\textbf{Dataset.}
For sparse-view geometry reconstruction, we construct a comprehensive benchmark on the ETH3D~\cite{schops2017multi}, DTU~\cite{jensen2014large}, and TUM~\cite{sturm12iros_ws} datasets, which feature diverse scenes including objects, indoor scenes, and outdoor scenes.
%

\noindent\textbf{Metrics and Baselines.}
We use the accuracy and completion metrics~\cite{jensen2014large} to assess point cloud quality. We compare our method against recent state-of-the-art methods like DUSt3R~\cite{wang2024dust3r}, MASt3R~\cite{leroy2024grounding}, and Spann3R~\cite{wang2024spann3r}.
We also include conventional SfM methods like COLMAP~\cite{schoenberger2016sfm} for comparison.
\begin{figure*}
\begin{center}
\vspace{-1em}
\includegraphics[width=\textwidth]{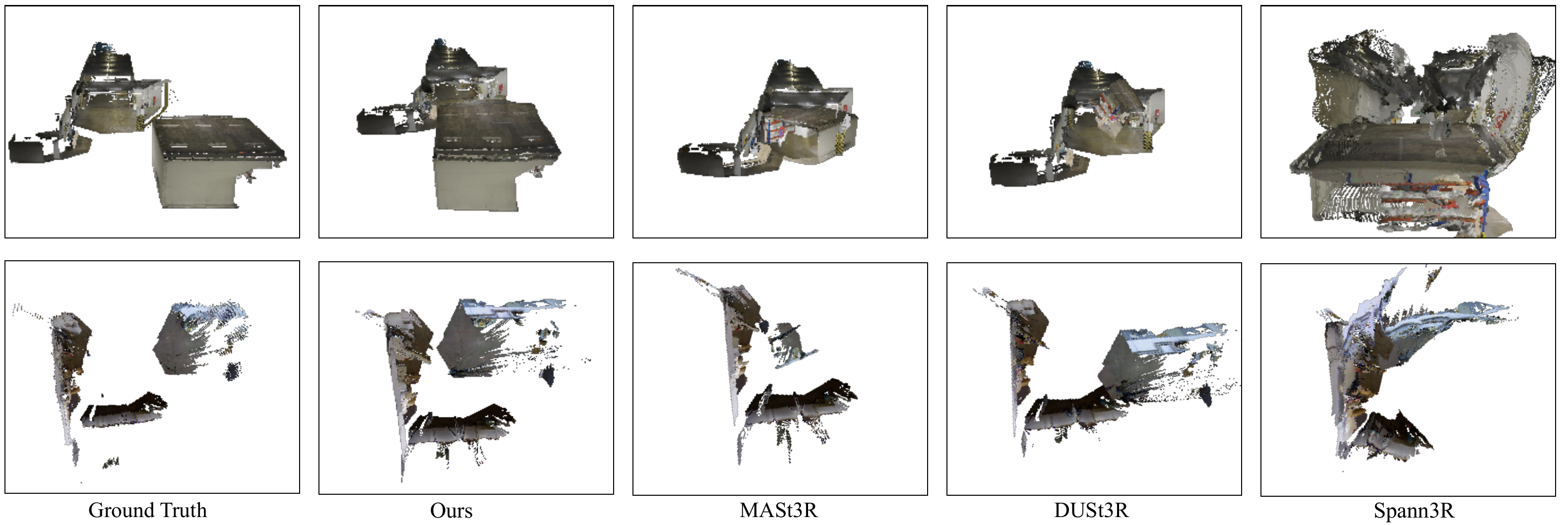}
\end{center}
\caption{
    \textbf{Qualitative Comparison results for sparse-view 3D reconstruction.}
    We visualize the 3D pointmaps of MASt3R~\cite{leroy2024grounding}, DUSt3R~\cite{wang2024dust3r}, Spann3R~\cite{wang2024spann3r} and our \method on ETH3D and TUM dataset.
}
\vspace{-1em}
\label{fig:reconstruction}
\end{figure*}

\noindent\textbf{Comparison.}
As shown in \cref{tab:recon}, our method achieves higher performance than optimization-based methods like DUSt3R~\cite{wang2024dust3r} and MASt3R~\cite{leroy2024grounding}, as well as recent feed-forward methods such as Spann3R~\cite{wang2024spann3r}.
Compared to DUSt3R and MASt3R, our method not only achieves better reconstruction quality but also offers faster inference, as our feed-forward reconstructor directly leverages multi-view information, avoiding their two-stage pipeline of two-view geometry estimation followed by global alignment post-processing.
Our method also outperforms Spann3R by leveraging camera poses as geometric priors in our two-stage geometry learning paradigm, rather than directly regressing point maps through neural networks.
We show qualitative comparisons in \cref{fig:reconstruction}. Our model achieves better geometry reconstruction with less noise compared to other baselines.


\subsection{Novel-view Synthesis}
\noindent\textbf{Dataset.} 
We evaluate rendering quality on two datasets: the RealEstate10K~\cite{zhou2018stereo}  and DL3DV~\cite{Ling_2024_CVPR}. For RealEstate10K, a widely used benchmark for novel view synthesis, we fine-tune our network using two-view images following the NopoSplat protocol~\cite{ye2024no}, adding intrinsic as a condition. Since we focus on sparse views, our RealEstate10K tests only the results from the NopoSplat test split, specifically in the overlapping regions with low and medium overlap.
For DL3DV, we sample 8 views for each testing video sequence as the input, and another 9 views as the groundtruth for evaluation. The sampling interval is randomly chosen, selected between 8 and 24. A total of 100 scenes are used as the test set.

\noindent\textbf{Metrics and Baselines.}
To evaluate rendering quality in the RealEstate10K~\cite{zhou2018stereo} and DL3DV~\cite{Ling_2024_CVPR} datasets, we employ PSNR, SSIM, and LPIPS metrics.
To evaluate rendering quality, we compare with pose-free methods, including CoPoNeRF~\cite{hong2023unifying}, Splatt3R, as well as pose-required methods, such as MVSplat~\cite{chen2024mvsplat} and PixelSplat~\cite{Charatan_2024_CVPR}.
We evaluate both MVSplat and pixelSplat using two-view inputs, selecting the two closest input views relative to the target view. Although MVSplat supports multiple input views, we find its performance degrades with additional views, as demonstrated in \supp. 
\begin{figure*}

\begin{center}
\includegraphics[width=\textwidth]{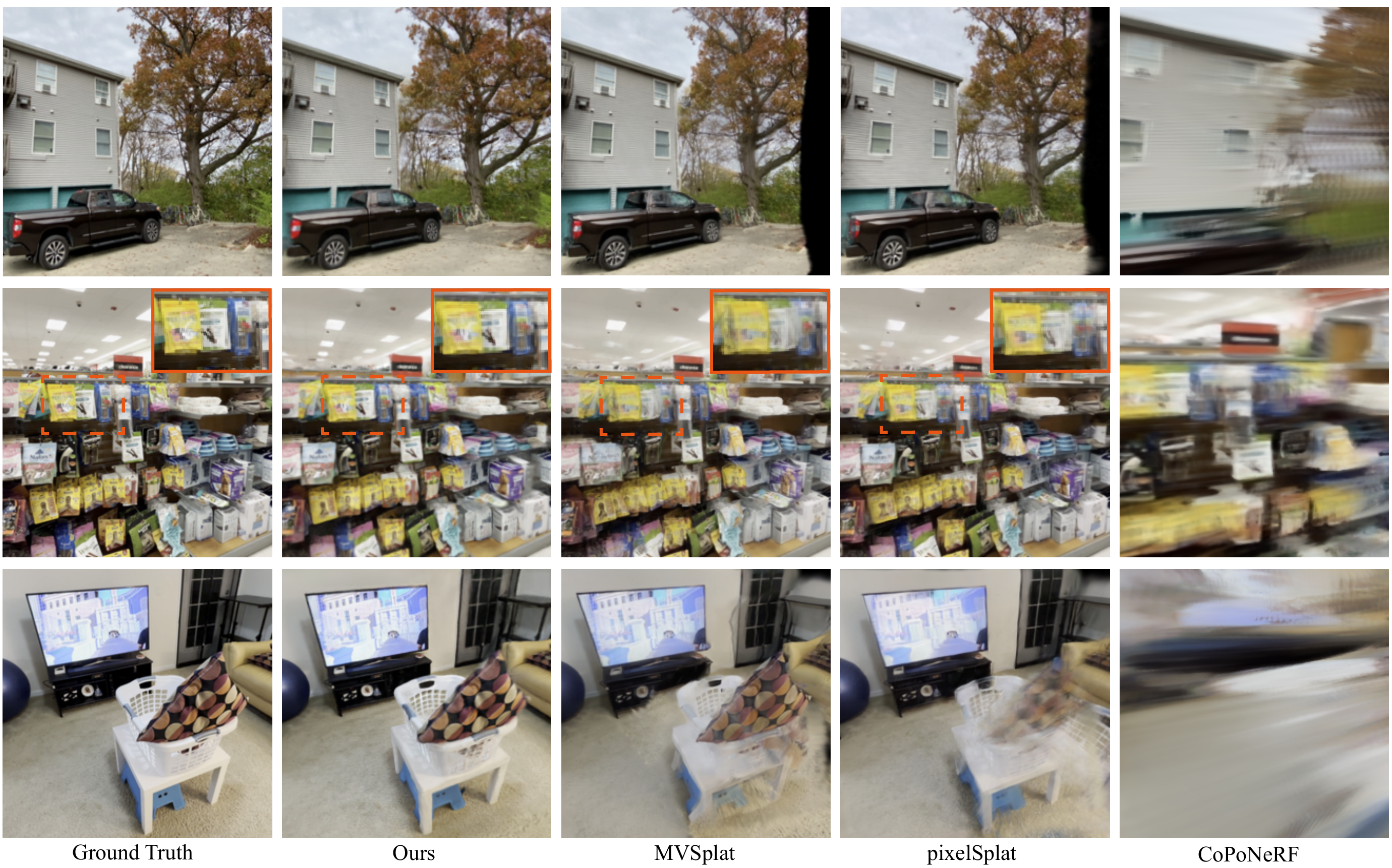}

\end{center}
\vspace{-1em}
\caption{
    \textbf{Qualitative Comparison results for novel-view synthesis.}
    We visualize the rendering results from the DL3DV dataset, which shows that our method obtains high-quality rendering from sparse-view, uncalibrated input images. 
}
\label{fig:rendering-comparison}
\vspace{-1em}

\end{figure*}


\begin{table}
\centering
\caption{\textbf{Comparison of novel view rendering on the DL3DV}.}
\small
\begin{tabular}{l| c c c}
\toprule
Method & PSNR$\uparrow$ & SSIM$\uparrow$ & LPIPS$\downarrow$ \\
\midrule
CoPoNeRF  & 16.06  & 0.472 & 0.474   \\
pixelSplat & 22.55  & 0.727 & 0.192   \\
MVSplat & 22.08 & 0.717 & 0.189 \\
Ours (2 views) & 23.04 & 0.725 & \textbf{0.182} \\ 
Ours (8 views) & \textbf{23.33} & \textbf{0.746} & 0.237  \\

\bottomrule
\end{tabular}
\vspace{-1em}
\label{comparison_results_realestate10k}
\label{tab:nvs-dl3dv}
\end{table}


\paragraph{Comparison.}
\cref{tab:nvs-dl3dv} and \cref{tab:nvs-realestate10k} report the quantitative evaluation results for novel-view synthesis on DL3DV and RealEstate10K, respectively. 
Our approach substantially outperforms pose-free methods like CoPoNeRF~\cite{hong2023unifying} and Splatt3R~\cite{smart2024splatt3r}, even better than the pose-required methods, including MVSplat~\cite{chen2024mvsplat} and PixelSplat~\cite{Charatan_2024_CVPR} that take camera poses provided by the dataset to achieve high-fidelity rendering. Our model doesn't require camera extrinsic information, which makes our method more applicable in real-world settings. 
\cref{fig:rendering-comparison} qualitatively illustrates that our pose-free method obtains higher rendering quality than the compared baselines.

\begin{table}
\centering
\caption{\textbf{Comparison of novel view rendering on the RealEstate10k.} We evaluate all methods using 2 views.}
\vspace{-0.5em}
\small
\begin{tabular}{l|ccc}
\toprule
Pose-required Method & PSNR↑ & SSIM↑ & LPIPS↓ \\
\midrule
pixelNeRF & 19.396 & 0.621 & 0.496 \\
AttnRend & 21.338 & 0.728 & 0.304 \\
pixelSplat & 22.495 & 0.777 & 0.210 \\
MVSplat & 22.568 & 0.781 & 0.200 \\
\midrule
Pose-free Method & PSNR↑ & SSIM↑ & LPIPS↓ \\
\midrule
Splatt3R & 15.113 & 0.492 & 0.442 \\
CoPoNeRF & 19.843 & 0.652 & 0.360 \\
Ours & \textbf{23.765} & \textbf{0.801} & \textbf{0.191} \\
\bottomrule
\end{tabular}
\vspace{-1em}
\label{tab:nvs-realestate10k}
\end{table}
\subsection{Ablation Study}
We conduct an ablation study to evaluate the effectiveness of each component in our method. For this study, we select the BlendedMVS~\cite{yao2020blendedmvs} dataset.
We randomly split the dataset of 95\% scenes in BlendedMVS dataset as the training set and keep the rest 5\% for testing.
Our ablations exclude rendering loss to focus on geometric results with various design choices.
``\textit{w/o} pose'' means our reconstructor is only conditioned on multiview images without inputting predicted camera poses. ``\textit{w/o} camera-centric'' means that we directly use a transformer with the same parameter size to output the global geometry without predicting camera-centric point maps. ``\textit{w/o} joint training'' means first training the camera predictor separately and then fixing it while training the reconstructor. 
"\textit{w/o} DPT head" denotes our ablation where we substitute the DPT decoder with a shallow MLP for point-map regression. "\textit{w/} rendering loss" refers to our configuration that incorporates  rendering loss during training to evaluate its impact on geometric accuracy. 

As demonstrated in \cref{tab:ablation-study}, using camera poses as proxies substantially improves geometry learning. Our two-stage approach with camera-centric geometry enhances performance, and the multi-task learning paradigm adds further improvements. DPT head plays a crucial role by explicitly accounting for spatial relationships during upsampling, whereas the shallow MLP lacks this consideration. The rendering loss has both positive and negative effects. While its dense supervision enhances COMP by supervising regions without ground truth point clouds, it slightly reduces ACC due to its lower accuracy compared to actual ground truth data.

\begin{table}
\centering
\caption{
\textbf{Ablation Study.} We evaluate the accuracy, completeness and the overall Chamfer distance on the testing set of BlendedMVS~\cite{yao2020blendedmvs} for the learned geometry. 
}
\small
\begin{tabular}{l|ccc}
\toprule
Method & ACC. $\downarrow$ & COMP. $\downarrow$ & Overall$\downarrow$ \\
\midrule
\textit{w/o} DPT head & 0.0399 & 0.0473 &  0.0436 \\
\textit{w/o} pose & 0.0283 & 0.0356 &  0.0319 \\
\textit{w/o} camera-centric  & 0.0265 & 0.0326 & 0.0295 \\
\textit{w/o} joint training  & 0.0276 & 0.0322 & 0.0299 \\
\textit{w/} rendering loss  & 0.0320 & \textbf{0.0244} & \textbf{0.0282}\\

\midrule
Ours  & \textbf{0.0250} & 0.0325 & 0.0288 \\
\bottomrule
\end{tabular}
\vspace{-1em}
\label{tab:ablation-study}
\end{table}

\section{Discussion}\label{sec::conclusion}

We introduce \method, a feed-forward model that can infer high-quality camera poses, geometry, and appearance from sparse-view uncalibrated images within 0.5 seconds.
We propose a novel cascade learning paradigm that progressively estimates camera poses, geometry, and appearance, leading to substantial improvements over previous methods.
Our model, trained on a set of public datasets, learns strong reconstruction priors that generalize robustly to challenging scenarios, such as very sparse views captured in real-world settings, enabling photo-realistic novel view synthesis.


\section*{Acknowledgements}
We gratefully acknowledge Tao Xu for his assistance with the evaluation, Yuanbo Xiangli for insightful discussions, and Xingyi He for his Blender visualization code.

{
    \small
    \bibliographystyle{ieeenat_fullname}
    \bibliography{reference}
}

\clearpage 
\newpage
\newpage 
\appendix

\section*{Appendix}

In this supplementary material, we include: (1) extended implementation details covering data processing and training strategies, (2) additional experimental results, and (3) a comprehensive description of the network architecture.

\section{Implementation Details}

\paragraph{Data Processing.} 
We follow the processing protocol of DUSt3R to generate point maps for most datasets. 
However, the DL3DV dataset only provides the annotation for  camera parameters.  
To include DL3DV into our training framework, we use the multi-view stereo algorithm from COLMAP to annotate per-frame depth maps, which are then converted into point maps. 
%
Additionally, we utilize multi-view photometric and geometric consistency to eliminate noisy depth~\cite{yan2020dense}.
%
%
%
%
%
For the datasets captured as video sequences, we randomly select $8$ images from a single video clip, with each video clip containing no more than 250 frames. 
For multi-view image datasets, we randomly select $8$ images per scene. 
%

\paragraph{Baselines for Novel View Synthesis.}
We compare our novel view synthesis results with MVSplat~\cite{chen2024mvsplat}, pixelSplat~\cite{charatan2024pixelsplat}, and CoPoNeRF~\cite{hong2023unifying} on the DL3DV dataset~\cite{Ling_2024_CVPR}.
%
%
However, these methods were originally trained on only $2$ views and perform not well under our sparse-view setting of $8$ views. To ensure a fair comparison, we selected the two source views closest to the target rendering view as inputs for these baselines (e.g., MVSplat). 
We found that selecting two closest source views significantly improved their rendering quality compared to using all $8$ views directly.
%




\paragraph{Numbers of Input Image.}
We have two camera latents: one for the first image (reference), and one is shared by all other images (source). The source token is duplicated N-1 times. Therefore, the model can process any number of input images.

\section{Experiments}
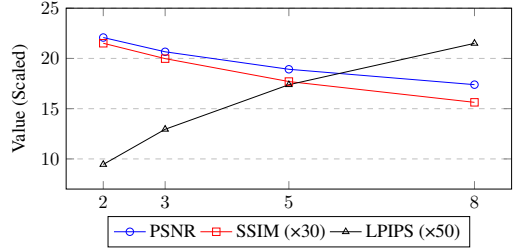
\begin{figure}
\centering
\begin{tikzpicture}[scale=0.7][hbtp]
    \begin{axis}[
        width=10cm,
        height=5cm,
        xlabel={Views},
        ylabel={Value (Scaled)},
        xtick={2,3,5,8},
        ymin=7, ymax=25,
        legend style={
            at={(0.5,-0.15)},
            anchor=north,
            legend columns=-1
        },
        ymajorgrids=true,
        grid style=dashed
    ]
    
    \addplot[
        color=blue,
        mark=o
    ]
    coordinates {
        (2,22.08)
        (3,20.66)
        (5,18.920)
        (8,17.39)
    };
    \addlegendentry{PSNR}

    \addplot[
        color=red,
        mark=square
    ]
    coordinates {
        (2,0.717*30)
        (3,0.666*30)
        (5,0.590*30)
        (8,0.521*30)
    };
    \addlegendentry{SSIM (×30)}

    \addplot[
        color=black,
        mark=triangle
    ]
    coordinates {
        (2,0.189*50)
        (3,0.259*50)
        (5,0.3477*50)
        (8,0.430*50)
    };
    \addlegendentry{LPIPS (×50)}
    \end{axis}
\end{tikzpicture}
\caption{\textbf{Relationship between MVSplat Performance and Input Views.}}
\label{view_number}
\end{figure}
\paragraph{Relationship between MVSplat Performance and Input Views.}
We evaluated MVSplat with two views because its
performance degrades with additional input frames, as shown in~\cref{view_number}, as
demonstrated in the figure above. We therefore reported
its optimal results.

\begin{table}[h]
\centering
\caption{\textbf{Performance with a Varying Number of Input Frames.} We study the impact of changing the number of input views on the performance of our method on the DTU dataset. }
\label{tab:ab_num}
\resizebox{0.48\textwidth }{!}{
\begin{tabular}{lcccccc}
\toprule
\textbf{Metric} & \textbf{2 Views} & \textbf{6 Views} & \textbf{10 Views} & \textbf{16 Views} & \textbf{25 Views} \\
\midrule
AUC@30° $\uparrow$ & 59.09 & 70.45 & 80.15 & 81.52 & \textbf{81.81} \\
ACC. $\downarrow$ & 4.07  & 2.79  &  0.94   &  \textbf{0.24}  &  0.30  \\
\bottomrule
\end{tabular}
}
\end{table}

\paragraph{Study between Performance and the Number of Frames.}
We analyzed the impact of varying the number of frames on pose and point map estimation using the DTU dataset.  
For this experiment, we randomly selected 2, 6, 10, 16, and 25 source views while fixing two query views for testing pose accuracy and for evaluating surface accuracy. Under the 2-view setting, our method generates a reasonable shape, but its precision remains limited. The results demonstrate that increasing the number of views leads to improvements in both pose and surface accuracy. However, these improvements gradually plateau as the number of views continues to grow, as shown in \cref{tab:ab_num}.

\begin{figure}
\begin{center}
\includegraphics[width=0.48\textwidth]{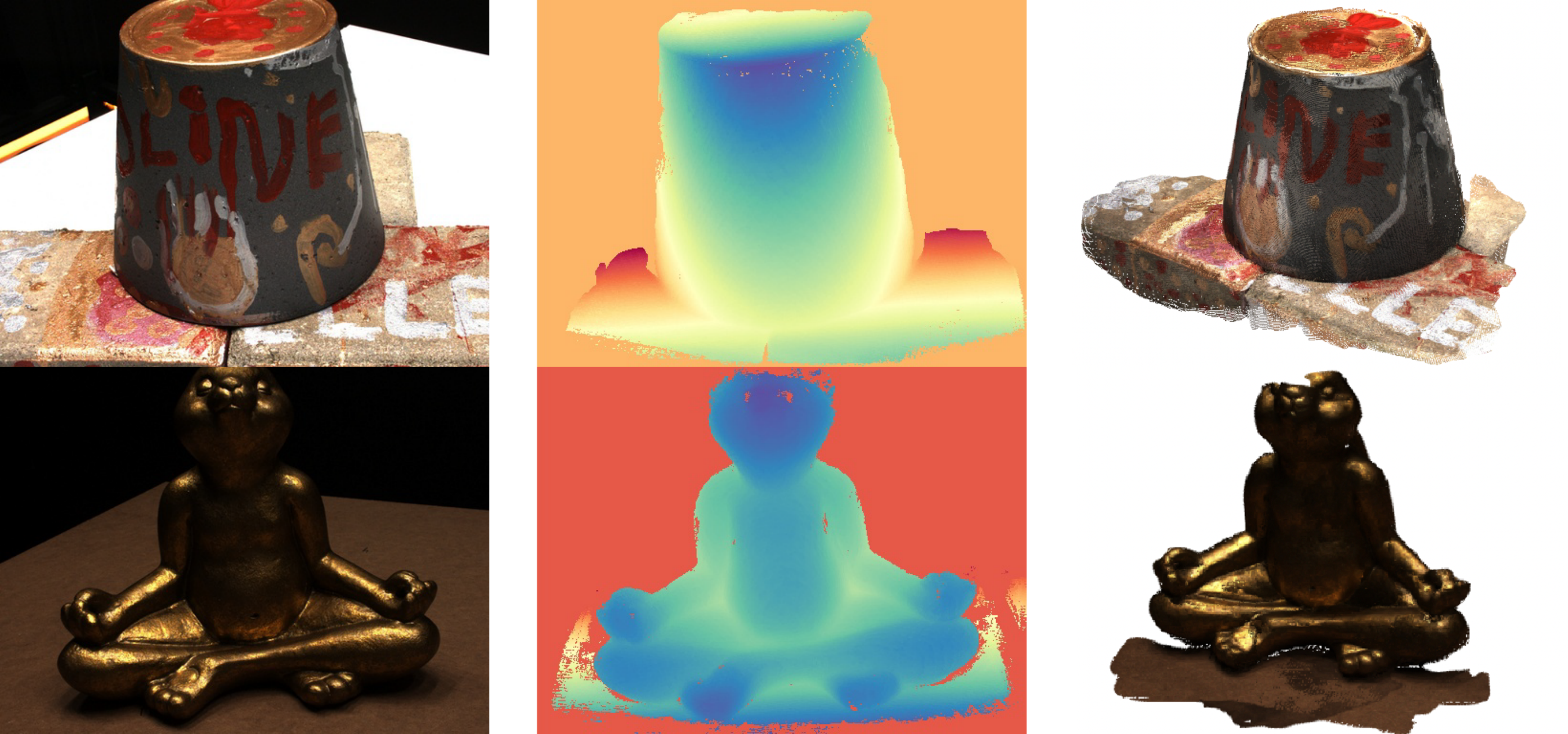}
\end{center}
\caption{
    \textbf{Qualitative Visualization of Sparse-view 3D Reconstruction on the DTU dataset.}
    We visualize the input image (left), depth map (middle), and point cloud (right).
}
\label{fig:depth}
\end{figure}

\paragraph{Dense View 3D Reconstruction on the DTU dataset.}
We present the results for dense view 3D reconstruction on the DTU dataset in \cref{dtu_dense}, although dense reconstruction is not our primary objective.
%
%
As observed, our method achieves better results compared to DUSt3R but falls short of MASt3R. 
This is expected since our approach is not tailored for dense reconstruction, whereas MASt3R is specifically optimized for it through the training of matching heads.  

\paragraph{Visualization of Sparse-view Reconstruction.} We present the visualizations of our sparse-view reconstruction results on the DTU dataset in \cref{fig:depth}.


\begin{table}
\centering
\caption{\textbf{Dense View 3D Reconstruction on the DTU dataset.} We compare our method with baseline approaches using accuracy, completeness, and overall metrics under the dense view setting.}
\resizebox{3in}{!}{
\begin{tabular}{lccc}
\toprule
      Methods &  Accuracy$\downarrow$ &  Completion$\downarrow$ &  Overall$\uparrow$ \\
\midrule
         Camp~\cite{camp} &     0.835 &       0.554 &    0.695 \\
         Furu~\cite{furu} &     0.613 &       0.941 &    0.777 \\
         Tola~\cite{tola} &     0.342 &       1.190 &    0.766 \\
       Gipuma~\cite{gipuma} &     0.283 &       0.873 &    0.578 \\
\midrule
       MVSNet~\cite{mvsnet} &     0.396 &       0.527 &    0.462 \\
   CVP-MVSNet~\cite{cvp-mvsnet} &     0.296 &       0.406 &    0.351 \\
      UCS-Net~\cite{ucs-net} &     0.338 &       0.349 &    0.447 \\
      CER-MVS~\cite{cermvs} &     0.359 &       0.305 &    0.332 \\
        CIDER~\cite{cider} &     0.417 &       0.437 &    0.427 \\
PatchmatchNet~\cite{pathcmatchnet} &     0.427 &       0.377 &    0.417 \\
    GeoMVSNet~\cite{geomvsnet} &     0.331 &       0.259 &    0.295 \\
    MASt3R~\cite{leroy2024grounding} &     0.403 &       0.344 &    0.374 \\
\midrule
       DUSt3R~\cite{wang2024dust3r} &     2.677 &       0.805 &    1.741 \\
       Ours &  \textbf{1.932} &  \textbf{0.715} & \textbf{1.321}\\

\bottomrule
\end{tabular}
}
\label{dtu_dense}
\end{table}



\end{document}